%% file: main.tex
\documentclass[twocolumn]{article}
\usepackage[a4paper,margin=1in]{geometry}
\usepackage[backend=biber]{biblatex}
\usepackage{graphicx}
\usepackage{algorithm}
\usepackage{algpseudocode}
\usepackage{amsmath, amsfonts}
\usepackage{hyperref}
\usepackage{csquotes}
\usepackage{parskip}
\usepackage{subfiles}
\usepackage{minted}

\DeclareMathOperator*{\opt}{opt}

\newcommand{\compact}[1]{
\begingroup
\setlength{\parskip}{0pt}
#1
\endgroup
}

\begin{document}

\title{Exploration--Exploitation--Evaluation (EEE): A Framework for Metaheuristic Algorithms in Combinatorial Optimization}
\author{Ethan Davis}
\date{}
\maketitle

\begin{abstract}
We introduce a framework for applying metaheuristic algorithms, such as ant colony optimization (ACO), to combinatorial optimization problems (COPs) like the traveling salesman problem (TSP). The framework consists of three sequential stages: broad exploration of the parameter space, exploitation of top-performing parameters, and uncertainty quantification (UQ) to assess the reliability of results. As a case study, we apply ACO to the TSPLIB berlin52 dataset, which has a known optimal tour length of 7542. Using our framework, we calculate that the probability of ACO finding the global optimum is approximately 1/40 in a single run and improves to 1/5 when aggregated over ten runs.
\end{abstract}


\section{Introduction}

\subfile{body/introduction}

\section{Methods}

\subfile{body/methods/eee}

\subfile{body/methods/case_study}

\subfile{body/methods/implementation_details}

\section{Results}

\subfile{body/results/exploration}

\subfile{body/results/exploitation}

\subfile{body/results/evaluation}

\subfile{body/results/software_engineering}

\section{Discussion}

\subfile{body/discussion/interpretation}

\subfile{body/discussion/strengths_limitations}

\subfile{body/discussion/future}

\section{Conclusion}

\subfile{body/conclusion}

\appendix

\subfile{appendix/optimization_metaheuristics}

\subfile{appendix/cluster}

\subfile{appendix/results}

\printbibliography

\end{document}

%% file: body/introduction.tex
We encounter combinatorial optimization problems (COPs) in daily life all the time. Take for example, the traveling salesman problem (TSP) that is a combinatorial optimization problem. The TSP can be simply defined as follows: Given $n$ cities and the distance between each pair of cities, seek a tour (a closed path) with minimum distance that visits each city in sequence once and only once and returns directly to the first city. This problem is seen, for example, by truck drivers whose job it is to make deliveries at multiple cities starting and ending at the center of operations, and seek the fastest route. As it turns out, the TSP is an NP-hard problem.

The total number of solutions of TSP is given by the formula $(n-1)!/2$. That is, the number of ways to visit all cities once, and return home, in an undirected path. Naive solutions such as exhaustive search (ES) are not capable of solving the TSP in a reasonable amount of time. Other algorithms, such as greedy algorithms (GA), have also been used to solve optimization problems. However, GA solutions are known to often have premature convergence. This motivates other algorithms for solving combinatorial optimization problems, and metaheuristic algorithms is a category of approaches that have been empirically successful at finding optimum solutions quickly \autocite{Tsai2023}.

There is a gap in standardization, though, for how to benchmark metaheuristic algorithms for the purpose of solving COPs, and continuous optimization problems. We care not only about the effectiveness of a metaheuristic algorithm to find optimum solutions, but also their consistency at doing so. This motivates the development of a unified framework that systematically evaluates both the exploratory and exploitative capabilities of metaheuristics while quantifying their reliability across runs \autocite{Tsai2023}.

We introduce the Exploration--Exploitation--Evaluation (EEE) framework for this purpose. In the \textit{Exploration} stage we perform a broad probe of the parameter space of a metaheuristic algorithm. For the \textit{Exploitation} stage, we move on to a deep search of the parameter space with top-performing parameters found in the previous stage. At the end of this stage, we are able to estimate of the success rate of a metaheuristic algorithm to optimally solve a COP. Then, the \textit{Evaluation} stage is used to evaluate the stability of the success rate we found.

In this paper, we describe our EEE framework in detail. We also present the results of a case study with our EEE framework that uses ant colony optimization (ACO) as a metaheuristic algorithm to solve the TSP. For our case study, we use the TSPLIB \textit{berlin52} dataset that has a known optimum solution of 7542 \autocite{Reinelt1991}. We implement ACO in Java and the Hadoop MapReduce framework, and our software is publicly available on GitHub \autocite{Davis2025}. For execution of our software, we create a Hadoop cluster of AWS EC2 instances, and details for how we made this cluster can be found in \autoref{Appendix:Cluster}. Our case study of the EEE framework finds that ACO reaches the optimum solution of TSP for berlin52 with probability 1/40 across one run, and improves to 1/5 across ten runs.

%% file: body/methods/eee.tex
\subsection{Exploration--Exploitation--Evaluation (EEE) Framework}

The Goal--Question--Metric (GQM) approach, and its extension GQMM (which incorporates models), provides a structured way to link abstract research goals with specific questions and measurable outcomes. To structure our methodology, we adopt the GQMM framework so that our research objectives are systematically linked to measurable outcomes. \autoref{Table:GQMM} presents an overview of this framework as applied to our study, outlining the overarching research goal, the guiding questions derived from it, and the associated metrics and models used for evaluation.

\begin{table}[h]
\centering
\caption{Summary of research}
\label{Table:GQMM}
\begin{tabular}{|l|l|}
\hline
\textbf{Goal} & Evaluate metaheuristics for COPs \\
\hline
\textbf{Question} & What are the optimal parameters? \\
\textbf{Metric} & Parameter ranking \\
\textbf{Method} & Sampling the parameter space \\
\hline
\textbf{Question} & How often is optimum reached? \\
\textbf{Metric} & Success rate \\
\textbf{Method} & Experimental runs \\
\hline
\textbf{Question} & How reliable is convergence? \\
\textbf{Metric} & Variance and standard deviation \\
\textbf{Method} & Replication \\
\hline
\end{tabular}
\end{table}

To find the optimal parameters for a metaheuristic algorithm, such that its convergence is fast and does not fall into local optima early, we perform repeated sampling from the parameter space. For example, the ACO metaheuristic algorithm parameters include the initial values $\tau$, the number of ants, the number of iterations for convergence, the Greek symbols $\alpha$, $\beta$, and $\rho$, and the constant $Q$. See \autoref{Appendix:Meta} for more details about the parameters of ACO. By defining a range for each parameter, a sampling method can be executed to explore the parameter space.

\begin{figure}[h]
\centering
\includegraphics[width=1.0\linewidth]{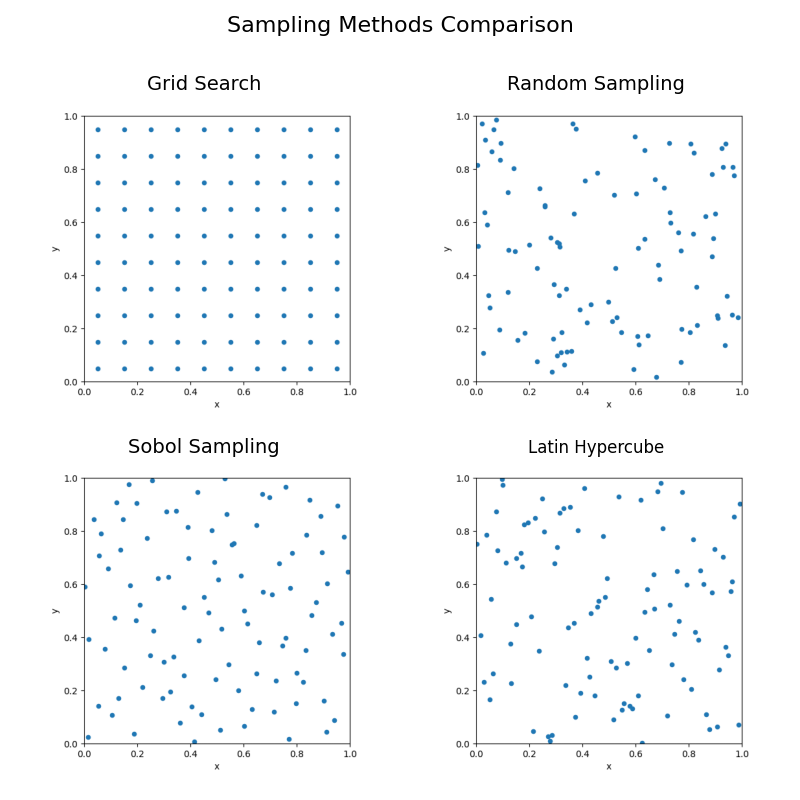}
\caption{Comparison of sampling methods}
\label{Figure:Sampling}
\end{figure}

Different sampling methods exist, and classical techniques are grid search and random sampling. Grid search covers a regular lattice of points that is deterministic and offers good coverage but scales poorly with dimensions. Random sampling is simple but can cluster and leave gaps. Quasi-random sampling methods are more effective. Sobol sampling generates low-discrepancy quasi-random points from deterministic sequences designed to fill a space as uniformly as possible. Latin hypercubes (LHS) are stratified so that each one-dimensional projection is well covered. \autoref{Figure:Sampling} shows 100 points sampled from a two-dimensional $[0,1] \times [0,1]$ range. Results from grid search and random sampling are sparse, however, Sobol sampling and LHS offer uniform coverage of the space.

After sampling tuples of parameters, for example $(\text{initial}\;\tau, \text{\# of ants}, \text{\# of iterations}, \alpha, \beta, \rho, Q)$ in the case of ACO, tuples are ranked from the results of exploratory runs. For example, for each parameter tuple $t_i$ where $i=1,2,\ldots,m$ and $m$ is the number of tuples, execute the metaheuristic algorithm $n$ number of times. When all runs have completed, the performance of parameter tuples can be ranked by the best solution from each run using summary statistics, such as $\operatorname{mean}$, $\operatorname{median}$, or $\operatorname{min}$. This summarizes the \textit{Exploration} stage of our EEE framework.

The \textit{Exploitation} stage from our EEE framework is to take the $p$ number of top-performing parameter tuples and rerun the metaheuristic algorithm with these tuples $q >> n$ number of times for the purpose of exploiting these parameters for good convergence in the solution space. For each of the $p$ number of parameter tuples, we collect the best solutions from each of the $q$ number of runs, and fit parametric probability distributions $f$ to the results using maximum likelihood estimation (MLE).

With metrics such as log likelihood (LL), Akaike information criterion (AIC), corrected AIC (AICc), and Bayesian information criterion (BIC), we can measure how well each parametric distribution fits to the best solutions generated from each parameter tuple. In general, information criterion metrics AIC, AICc, and BIC are regularized versions of LL, AICc is used for small samples, and BIC penalizes model complexity more strongly than AIC leading to simpler models. The formula for these measurements of maximum likelihood fit are as follows,

\compact{
\begin{align}
\operatorname{LL} &= \sum_{i=1}^{q}\ln{f(x_i \mid \theta)}, \\
\operatorname{AIC} &= 2k - 2\ln(\operatorname{LL}), \\
\operatorname{AICc} &= \operatorname{AIC} + \frac{2k(k+1)}{q-k-1}, \\
\operatorname{BIC} &= \ln(q)k - 2\ln(\operatorname{LL}),
\end{align}
}

where $q$ is the number of metaheuristic algorithm runs with independent and identically distributed (IID) results, that is, the sample size, and $k$ is the number of parameters used in the probability distribution.

We use these information criterion metrics to rank the parametric distributions by how well they fit the best solutions from runs of each parameter tuple and select the single top-performing distribution. With this distribution, we calculate the probability of reaching the optimum solution by using the cumulative distribution function (CDF). In summary, this methodology exploits the top-performing parameter tuples from exploration, fits parametric distributions to the best solutions from these exploited runs, and computes how often the optimum solution is reached by the top-performing distribution through the CDF.

The last stage of our EEE framework is \textit{Evaluation}, and in this stage we determine the consistency of results. We use repeated sampling methods of our empirical data to find the central tendency and spread of our results. For example, given bootstrapped samples from our data, we repeat steps from the previous stage of our EEE framework: We fit a parametric distribution to the bootstrapped data, and compute a probability by the CDF. Repeating this process many times gives us a distribution of probabilities that let us compute, for example, the mean and variance.

This summarizes our Exploration--Exploitation--Evaluation framework that can be used to evaluate the effectiveness and stability of metaheuristic algorithms for combinatorial optimization problems. Next, we describe a case study where we apply our EEE framework to evaluate ant colony optimization for the traveling salesman problem. The ACO metaheuristic algorithm and TSP COP are widely known, see \autoref{Appendix:Meta} for more details about them. Later, we discuss the strengths and limitations of the EEE framework, and the results of our case study.

%% file: body/methods/case_study.tex
\subsection{Case Study: ACO for TSP}

In \autoref{Appendix:Meta}, we define the ant colony optimization metaheuristic algorithm and the traveling salesman problem, see this appendix for a detailed background. For this case study, we use the TSPLIB \textit{berlin52} dataset \autocite{Renardy2021}. TSPLIB is a well-known repository of TSP datasets, and berlin52 is a specific example of a widely used TSPLIB dataset for benchmarking. The known optimum shortest path solution of berlin52 is 7544.36 when measured in Euclidean distance. However, TSPLIB records the optimum solution as 7542 as measured by its standard of integer edge length.

\begin{figure}[h]
\centering
\includegraphics[width=1.0\linewidth]{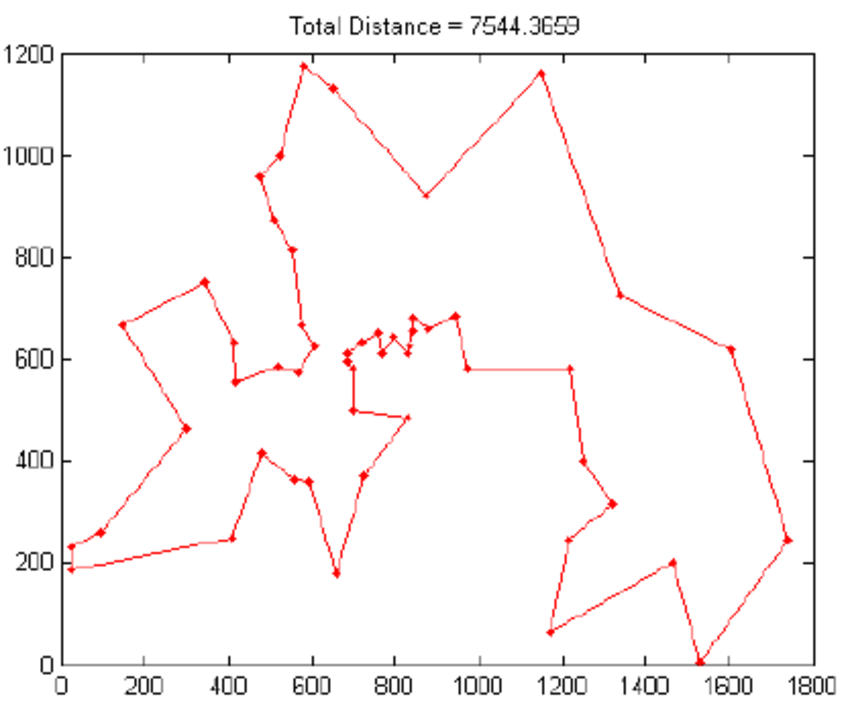}
\caption{Optimum path of berlin52}
\label{Figure:Sampling}
\end{figure}

The first stage of the EEE framework is \textit{Exploration}. We must scope the parameter space of our metaheuristic algorithm and choose a sampling method for drawing tuples of parameters from it. The parameters of ACO are the initial value of $\tau$, the number of ants, the number of iterations for convergence, the Greek symbols $\alpha$, $\beta$, and $\rho$, and the constant $Q$. To reduce the number of dimensions of the the parameter space and risk of sparse sampling, we make some parameters constant. We set all initial values $\tau_{i,j}$ equal to 1 which is the number of pheromones on each edge, the number of iterations for convergence equal to 10, and the constant $Q$ to 1000 which is proportional to the tour length for an ant.

Since we set all initial values $\tau_{i,j}$ equal to 1, we remove bias from the very beginning of convergence. State transitions from early iterations of ACO should be based on a priori beliefs, namely, the local information or distance associated with a certain state transition. After multiple iterations, transitions made by many ants build up a posteriori beliefs, that is, global information or pheromones associated with state transitions.

The number of iterations for convergence is controlled in our experiment. The \textit{Exploration} stage of our EEE framework is designed to broadly search the solution space with many tuples of parameters. Therefore, to reduce computation costs, we limit the number of iterations for convergence to be equal to 10 for all runs. Furthermore, for the purpose of ranking parameter tuples by summary statistics of their performance across a small and exploratory number of runs, we execute ACO 3 number of times for each parameter tuple.

Lastly, we choose the constant $Q$ to be 1000 because this is proportional to the tour length of an ant. We know this value of $Q$ is proportional to the tour length because we know the optimum tour length of berlin52 is 7542. We desire $Q$ to be proportional to the tour length because the ratio $\frac{Q}{L_k}$, where $L_k$ is the tour length of the $k$th ant, is used for pheromone updates.

The remaining parameters of ACO that we leave variable are the number of ants, and the Greek symbols $\alpha$, $\beta$, and $\rho$. The symbol $\alpha$ controls the importance of pheromone information, that is, it influences diversification of solutions. On the other hand, $\beta$ controls the importance of distance information, in other words, it drives intensification of solutions. Both these symbols create a balance of exploration and exploitation themselves that is separate from the \textit{Exploitation} and \textit{Exploration} stages of our EEE framework. Finding the optimal balance of these parameters is not intuitive.

The Greek symbol $\rho$ adds additional complexity to the balance of symbols $\alpha$ and $\beta$. The value of $\rho$ controls the pheromone evaporation rate. In other words, the environment in which parameters $\alpha$ and $\beta$ influence navigation of ants is adjusted each iteration of convergence by the parameter $\rho$. A high pheromone evaporation rate means a slower build up of pheromones, and therefore, a slower build of a posteriori information for decision making.

The last parameter that we keep variable in our experiment is the number of ants. This parameter influences the rate of information gain per iteration. With fewer ants, each iteration yields less information, so convergence is slower but computational cost per iteration is lower. On the other hand, too many ants can introduce noise or premature convergence if the algorithm amplifies mediocre solutions too quickly. None of the parameters, that is, $\alpha$, $\beta$, $\rho$, or the number of ants, is intuitive to tune, making them good candidates to keep variable in the parameter space of our experiment.

ACO has the following theoretical constraints on the Greek symbols $\alpha$, $\beta$, and $\rho$, and the number of ants: $\alpha > 0$, $\beta > 0$, $0 \leq \rho \leq 1$, and $\text{the number of ants}\;\geq1$. For our case study, we use the following ranges for these parameters:

\begin{align}
\alpha &\in [0.5, 2.0] \\
\beta &\in [1.0, 5.0] \\
\rho &\in [0.1, 0.9] \\
\text{ants} &\in [50, 250].
\end{align}

Our reason for defining the ranges of parameters as we did is based on empirical experience. These ranges offer moderate constraints on the parameters. Relaxing constraints beyond these ranges would risk ACO runs with premature stagnation, excessive randomness, and wasted computation.

To sample the parameter space for tuples of $\alpha$, $\beta$, $\rho$, and the number of ants, we use Sobol sampling. Grid search and random sampling sparsely cover the sampling space offering worse performance than pseudo-random sampling methods like Sobol sampling and Latin hypercubes. The literature has shown that Sobol sampling offers better coverage of the sampling space than LHS \autocite{Renardy2021}. There are multiple types of Sobol sampling, and for our case study, we use Saltelli's extension of the Sobol sequence. We have $D=4$ dimensions ($\alpha$, $\beta$, $\rho$, and the number of ants) and generate $N=8$ samples, altogether creating a total of $N(D+2)=48$ parameter tuples. A detailed explanation of Sobol sampling is beyond the scope of this paper.

In closing of the \textit{Exploration} stage of our EEE framework, where we execute ACO 3 number of times for each parameter tuple, we rank tuples by their mean shortest distance. For our case study, we take the 5 top-performing parameter tuples. These are used to execute ACO again for the \textit{Exploration} stage of our EEE framework. In this subsequent stage, we perform more iterations for convergence for all runs of ACO, and more runs for each parameter tuple. We keep the same constraints for the initial values of $\tau$, the number of ants, the Greek symbols $\alpha$, $\beta$, and $\rho$, and the constant $Q$. However, we change the number of iterations for convergence to be 30, and the number of runs per parameter tuple to be 10, for our case study.

In the \textit{Exploitation} stage, rather than ranking parameter tuples by their mean shortest distance across multiple runs, we rank them by their probability of reaching the optimum shortest distance. For our case study, we use Normal, Log-normal, Gamma, and Weibull distributions to fit to our data by MLE for each parameter tuple. Given these distributions, we rank distributions by AICc per parameter tuple. We use AICc rather than another information criterion because 10 data points per parameter tuple is a small dataset. Using the top-performing distribution per parameter tuple, we use the CDF to calculate $P(X\leq7542)$, the probability of reaching the optimum solution. This probability is how we rank the single top-performing parameter set.

At this point, we are ready for the \textit{Evaluation} stage of our EEE framework. We use bootstrap sampling and confidence intervals to measure the stability of our results. For $10,000$ iterations, we sample our empirical dataset with replacement 10 times. For each bootstrapped dataset, we use MLE to fit the top-performing distribution from the previous stage of the EEE framework, and we calculate the probability of the optimum solution by the CDF. The aggregate of these probabilities is used to calculate the confidence intervals. We report the mean probability and 95\% confidence intervals as the reliability of reaching the optimum solution.

%% file: body/methods/implementation_details.tex
\subsection{Implementation Details}

Our initial step before execution of ACO is the generate parameter tuples from Sobol sampling. For this purpose, we use the Python package, SALib \autocite{Iwanaga2022, Herman2017}. This package implements different types of Sobol sampling, and its interface makes it easy to encode the parameter space, and generate and write parameter tuples to a file. For both the \textit{Exploration} and \textit{Exploitation} stages of our EEE framework, we use CSV files encoding parameter tuples to automate runs of our ACO executable as follows.

\begin{minted}{ini}
index,alpha,beta,rho,ants
0,0.584,1.993,0.362,222
1,1.149,1.993,0.362,222
2,0.584,1.352,0.362,222
...
[number of tuples]
\end{minted}

We use Java and Hadoop MapReduce to implement ACO, and a cluster of AWS EC2 instances for execution. In this section we describe the details of our software design for ACO. Our Java and Hadoop software is stored in a public GitHub repository \cite{Davis2025}. For our case study, we used 4 networked instances of $\operatorname{t3.xlarge}$ EC2 instance types with 4 vCPUs and 16 GiB of memory each. Details for how we built a cluster of AWS EC2 instances can be found in \autoref{Appendix:Cluster}.

As input, our ACO executable accepts values for all parameters of ACO as defined by the client, namely, a human or an automated script. The pheromone table with initial values of $\tau$ is stored in a file, and the file path is passed as input. The format of the pheromone table is defined as follows,

\begin{minted}{ini}
source,destination,distance,pheromones
0,1,666,1
0,2,281,1
0,3,396,1
...
[number of edges]
\end{minted}

where each file line describes edge-centric information. For our software design, we had to make a tradeoff between passing the whole pheromone table as configurations between MapReduce operations, or to cache the pheromone table on the disk of each node in our cluster. Both designs are undesirable, in the former we pay for high cost network communications, and in the latter we pay for disk access. Our design chooses the latter option under the belief that the cost of local disk access is lower than network communications for the pheromone table.

There is also an ants file, whose file path is passed as input, and its contents is distributed across worker nodes. Each line of the ants file is an arbitrary value, however, the number of file lines encodes the number of ants for ACO. For example, we use the following format for the ants file.

\begin{minted}{ini}
ant
1
1
1
...
[number of ants]
\end{minted}

For each iteration of ACO convergence, a MapReduce job is run. When finished, each MapReduce job writes multiple files as output: The updated pheromone table, and the shortest distance from that iteration. The shortest distance is also written with an encoded string of the shortest path. If there are more iterations for convergence, the next iteration will read the updated pheromone table for operations. Also, between iterations, a helper operation will read the shortest distance information, and write a log entry if a new shortest distance was found.

\begin{figure}[h]
\centering
\includegraphics[width=1.0\linewidth]{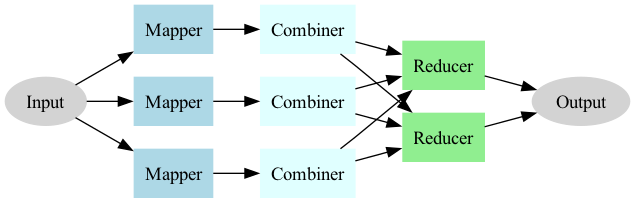}
\caption{Summary of a MapReduce job.}
\end{figure}

Inside each MapReduce job, the master node distributes work to all worker nodes, including itself. We use $\operatorname{Mapper}$, $\operatorname{Combiner}$, and $\operatorname{Reducer}$ operators. For each line of the ants file, the $\operatorname{Mapper}$ operator executes $\operatorname{map}()$. That is, each run of $\operatorname{map}()$ corresponds to the operations of one ant. Each ant is responsible for construction of one tour that is a candidate solution.

Solution construction is encapsulated by a plain old Java object (POJO) that is independent of Hadoop packages in its imports. In terms of software engineering SOLID principles \autocite{Martin2017}, this follows the dependency inversion principle (DIP) and single responsibility principle (SRP), which reduces the measurement of lack of cohesion of methods (LCOM) \autocite{McConnell2004}. This POJO is loosely coupled with the Hadoop framework, which is ideal in case we are also interested in using the Spark or message passing interface (MPI) frameworks, for example.

The output of $\operatorname{map}()$ is the candidate solution distance and string encoded path. Additionally, the $\operatorname{map}()$ operation outputs the distance (weight) and pheromone information. Later, in the $\operatorname{reduce{}}$ operation, this information will be aggregated from all ants into an updated pheromone table that subsequent iterations of ACO convergence use. The separate outputs from $\operatorname{map}()$ are identified by different keys so that the shuffle operation of MapReduce directs all output with the same key to the same worker nodes for the $\operatorname{reduce}()$ operation.

Before information is distributed from the $\operatorname{Mapper}$ operator to the $\operatorname{Reducer}$ operator, it is sent to a $\operatorname{Combiner}$ operator. This additional $\operatorname{Combiner}$ operator is optional to the MapReduce framework, but can greatly decrease network communication from information being sent from $\operatorname{Mapper}$ to $\operatorname{Reducer}$. The $\operatorname{Combiner}$ acts as an intermediate $\operatorname{Reducer}$, that aggregates information with the same key on the same worker node, before it is distributed via network communication for the shuffle operation that must direct all information with the same keys to the same worker nodes for aggregation. A $\operatorname{Combiner}$, that is an intermediate $\operatorname{Reducer}$, decreases the amount of network communication between worker nodes.

The aggregation performed by the $\operatorname{Reducer}$, and $\operatorname{Combiner}$, is to find the $\min$ shortest distance ant tour, and the $\operatorname{sum}$ of pheromone deltas, for all ants. The $\operatorname{Reducer}$ also performs pheromone evaporation, and assigns the final edge-centric pheromone update. At the end of the $\operatorname{reduce}()$ operation, these results are written to separate files, to be processed by the master node. This completes one iteration from ACO convergence.

For post-processing of our ACO results, we use Python and regular expressions to parse our log files, and build an in-memory object of best solutions. Using Matplotlib, NumPy, and SciPy \autocite{Harris2020, Hunter2007, Virtanen2020}, we create plots and table from summary statistics of our empirical best solutions data. The plots and tables offer a qualitative view of our results, and these summary statistics contribute to quantitative, data driven decision making when moving between different stages of our EEE framework.

%% file: body/results/exploration.tex
\subsection{Parameter Exploration}\label{Results:Exploration}

Sobol sampling generated 48 tuples of parameters. \autoref{Figure:Sobol} in \autoref{Appendix:Results} shows the sequence that was generated. For each of these run indexes, that is tuple of parameters, we execute the ant colony optimization metaheuristic algorithm to solve the traveling salesman problem for TSPLIB berlin52 a total of 3 number of times.

From the shortest distance results per parameter tuples, we create box plots to understand the quality of our results from executing ACO for 48 parameter tuples across 3 runs each. \autoref{Figure:BoxplotsExploration} from \autoref{Appendix:Results} shows these box plots, where it can be seen that many parameter tuples have similar performance, but some have a wider variance and some clearly have a worse average performance.

We use the mean and median of shortest distance per parameter tuple to quantitatively compare results, as can be seen in \autoref{Figure:TableExploration} in \autoref{Appendix:Results}. The 10 top-performing parameter tuples, that are identified by their run index, from both mean and median categories are very similar with 8/10 runs appearing in both categories. For the purpose of ranking the 5 top-performing parameter tuples for the \textit{Exploration} stage of our EEE framework, we use the mean.

Through a trial run, we found that the time needed to execute ACO for 10 iterations of convergence took approximately 4 minutes. Therefore, we calculated the time needed to execute 3 runs each for 48 parameter tuples would take $4 \times 3 \times 48 \div 60 = 9.6 \; \text{hours}$. In other words, approximately one period of execution over night. Our estimated runtime was correct, and the cost of execute was 10 USD.

%% file: body/results/exploitation.tex
\subsection{Exploitation Stage}\label{Results:Exploitation}

\begin{figure}[h]
\centering
\includegraphics[width=\linewidth]{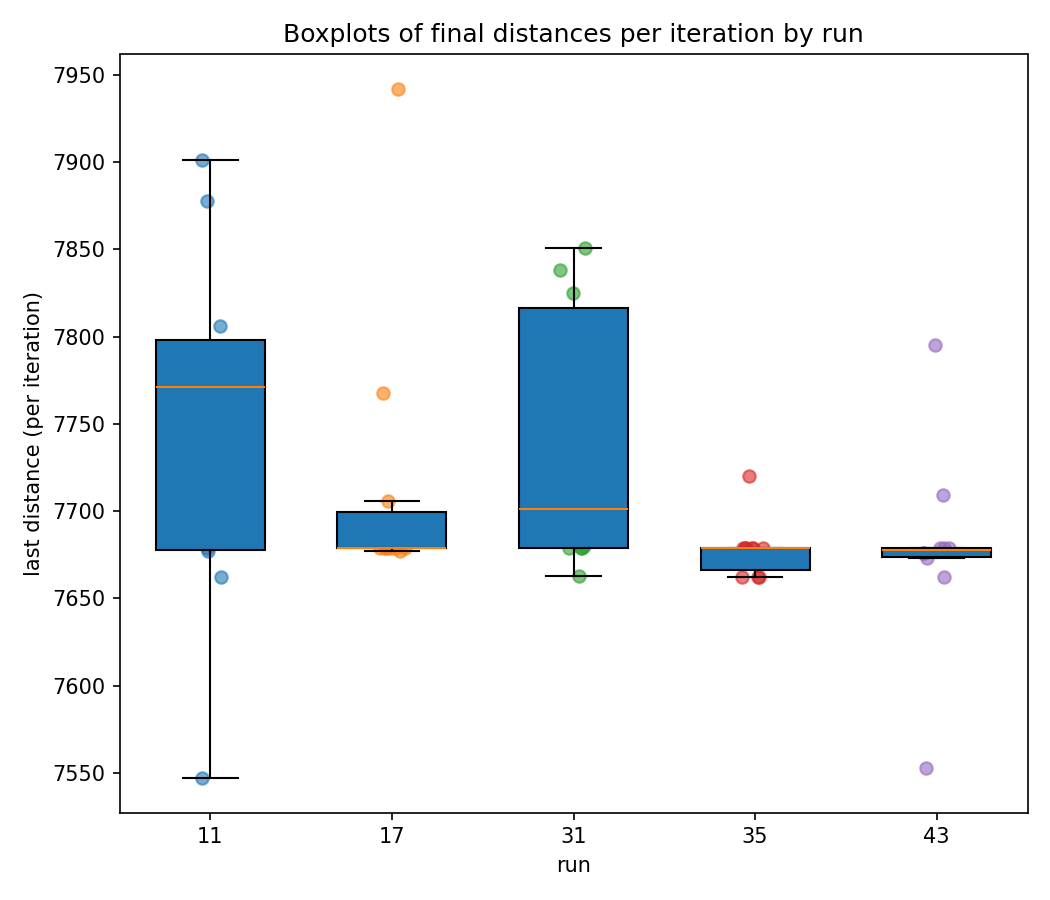}
\caption{Shortest distance box plots from the \textit{Exploitation} stage in our EEE framework.}
\label{Figure:BoxplotsExploitation}
\end{figure}

Given the 5 top-performing parameter tuples from the \textit{Exploitation} stage, we reran ACO with 30 iterations for convergence and 10 number of times per parameter tuple. \autoref{Figure:BoxplotsExploitation} shows the box plots of our shortest distance results from running ACO in our \textit{Exploitation} stage. It is interesting to note that mostly consistent runs of parameter tuples with low variance in their shortest distances, namely, runs with indexes 17, 35, and 43, neither come especially close to the optimum solution nor swing far in the opposite direction. On the other hand, runs with a wide variance, namely the run indexed by 11, swing from good to bad solutions.

\begin{figure}[h]
\centering
\includegraphics[width=\linewidth]{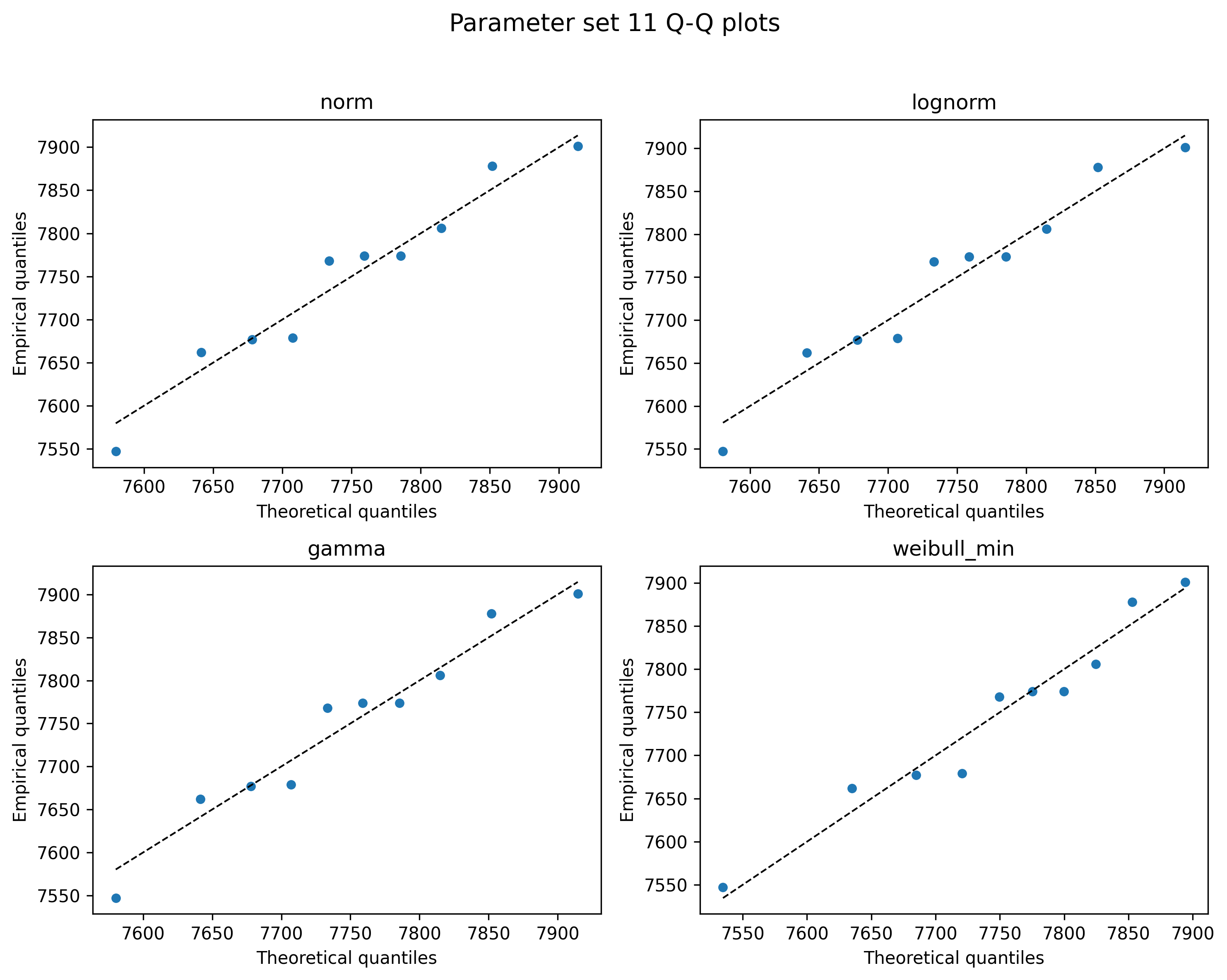}
\caption{Q-Q plots from runs of the parameter tuple with index 11.}
\label{Figure:QQPlots11}
\end{figure}

To rank these runs, we first fit Normal, Log-normal, Gamma, and Weibull distributions to each by the MLE. For a qualitative perspective of how well these distributions fit to our data, we create Q-Q plots. We find that the run indexed by parameter tuple 11 most closely, visually fits these distributions. \autoref{Figure:QQPlots11} shows our Q-Q plots for the run with index 11, and \autoref{Figure:QQPlotsExploitation} in \autoref{Appendix:Results} shows the Q-Q plots of the remaining parameter tuple runs. We emphasize that these plots do not tell us which parameter tuple has the best performance at finding the optimum shortest path, but rather how well our chosen parametric probability distributions fit to our data. Given the best fitting probability distribution of each parameter tuple, we rank parameter tuples by their probability of finding the optimum solution.

\begin{figure}[h]
\centering
\includegraphics[width=\linewidth]{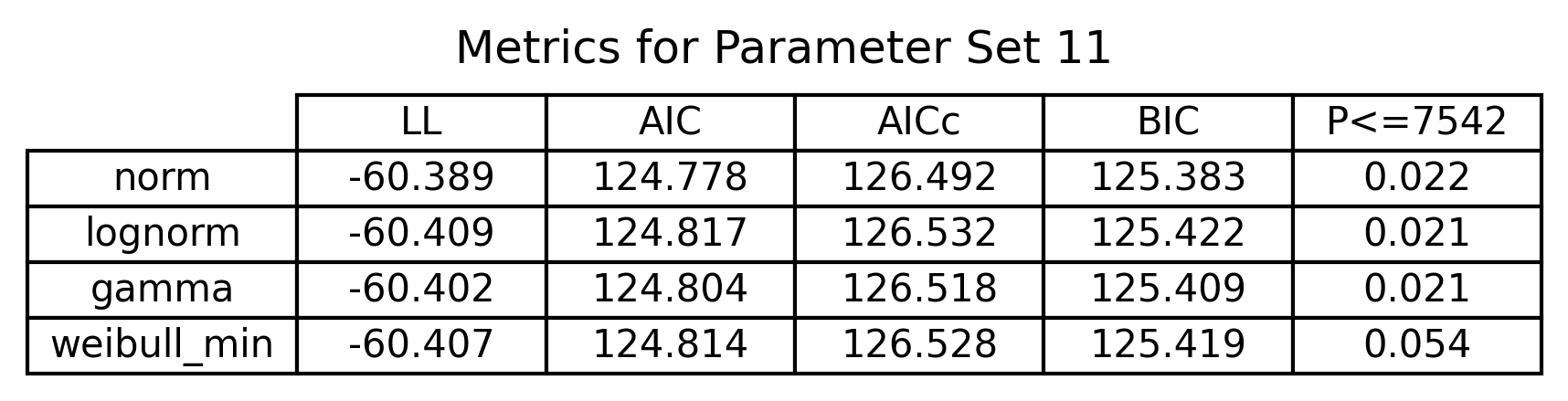}
\caption{Information criterion metrics from runs of the parameter tuple with index 11.}
\label{Figure:Metrics11}
\end{figure}

We calculate information criterion metrics to quantitatively compare how well each of our chosen parametric probability distributions, namely, Normal, Log-normal, Gamma, and Weibull, fit to our data. We rank distribution fit by the AICc, and then rank parameter tuples by the CDF calculated probability of a run reaching the optimum shortest path. By coincidence, parameter tuple with index 11 not only has the best fitting parametric distributions from Q-Q plots, it also has the highest probability of reaching the global optimum, as calculated by the CDF $P(X\leq7542)$ of the Normal distribution. \autoref{Figure:Metrics11} shows the information criterion metrics for parameter tuple with index 11, and its corresponding probability of reaching the global optimum across one run as approximately 1/40. \autoref{Figure:MetricsExploitation} in \autoref{Appendix:Results} shows the information criterion metrics from the other parameter tuples. Note that these results can be read by finding the minimum AICc in a given table and then reading its corresponding probability $P(X\leq7542)$.

In summary, we have completed the \textit{Exploration} and \textit{Exploitation} stages of our EEE framework and estimated that the probability of ACO reaching the optimum solution of TSPLIB berlin52 across one run to be 1/40. Now, we are interested in the reliability of our findings. For that purpose, we move on to the final stage of our EEE framework: \textit{Evaluation}.

Finally, we used our estimation from empirical data of ACO trial runs to estimate the cost of executing runs of our \textit{Exploitation} stage. Given that 10 iterations for ACO convergence takes roughly 4 minutes, and we planned to execute ACO with 30 iterations for convergence 10 number of times for each of our 5 top-performing parameter tuples, then runtime would take $4 \times 3 \times 10 \times 5 \div 60 = 10 \; \text{hours}$. In other words, roughly equal to the time needed for runtime of our \textit{Exploration} stage. Our estimations were correct, and cloud compute cost was 10 USD.

%% file: body/results/evaluation.tex
\subsection{Reliability Evaluation}

We used bootstrap sampling of our data for the purpose of uncertainty quantification. The data from runs of each parameter tuple was sampled with replacement 10 times for $10,000$ iterations, each bootstrapped dataset was fit with MLE using the best fitting parametric distribution from the \textit{Exploitation} stage, and the probability $P(X\leq7542)$ was calculated by the CDF. This gave us datasets for which we could calculate the mean and 95\% confidence intervals.

As in the \textit{Exploitation} stage, the parameter tuple with index 11 was also the single top-performing parameter tuple in our \textit{Evaluation} stage. We estimated that, across one run, on average, parameter tuple with index 11 reaches the optimum shortest distance with probability 1/40. Moreover, if we repeated our experiment many times, then, across one run, 95\% of the probabilities of reaching the optimum solution would come between 0 and above 9/100. Furthermore, across ten runs, the average becomes 1/5, and 95\% of the probabilities come between near zero and above 6/10. This wide range of probabilities could be attributed to the wide variance of performance from parameter tuple with index 11 as seen in \autoref{Figure:BoxplotsExploitation}.

\begin{figure}[h]
\centering
\includegraphics[width=\linewidth]{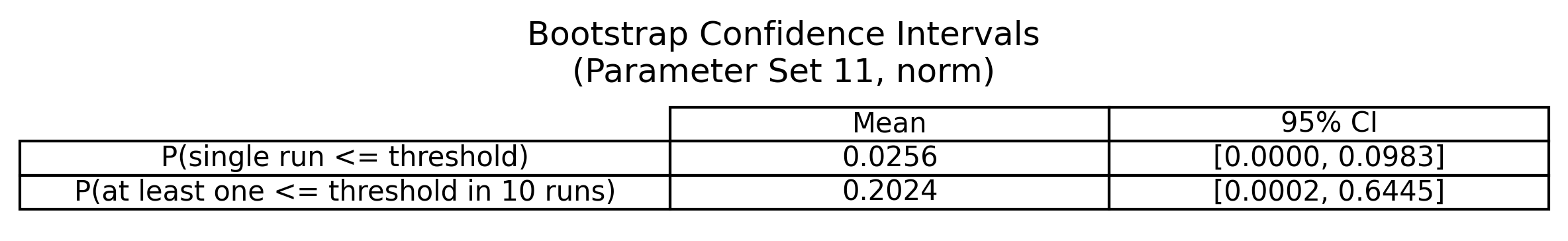}
\caption{Bootstrap confidence intervals from our \textit{Exploitation} stage of the EEE framework.}
\label{Figure:Bootstrap11}
\end{figure}

\autoref{Figure:Bootstrap11} shows the results of our uncertainty quantification for the \textit{Evaluation} stage of our EEE framework from runs of parameter tuple with index 11. Our remaining uncertainty quantification results are shown in \autoref{Figure:BootstrapEvaluation} that can be found in \autoref{Appendix:Results}. These results illustrate the variability in performance estimates and the confidence bounds obtained from repeated resampling.

%% file: body/results/software_engineering.tex
\subsection{Software Engineering}\label{Results:SoftwareEngineering}

In this section, we present quantitative measurements from our Java and Hadoop MapReduce implementation of ant colony optimization. These metrics can be used to indicate, for example, how readable our Java software is, how writable the Hadoop MapReduce framework is, and how modular our design is. The metrics we use to measure our ACO implementation are lines of code (LoC), lines of boilerplate code, cyclomatic complexity, and lack of cohesion of methods (LCOM).

Measuring LoC is simply calculated as the total number of source lines across all executable logic \autocite{McConnell2004}. The lines of boilerplate code are less than or equal to the LoC and indicate how much of the code base is devoted to administrative or structural tasks rather than business logic \autocite{McConnell2004}. While LoC provides a rough sense of system size or effort, it does not directly reflect code quality or complexity. The other metrics, described below, have more nuanced definitions that capture different structural and behavioral aspect of software.

Cyclomatic complexity measures the number of decision points in logic \autocite{McConnell2004}. For example, zero conditional statements in a sequence of commands, that is, a flat program, means that cyclomatic complexity equals 0. A program with one conditional statement means cyclomatic complexity is 1. Two separate, unnested conditional statements in-sequence means that cyclomatic complexity is still 1. Two conditional statements, one nested inside the other, makes cyclomatic complexity equal to 2. \autoref{Table:Complexity} summarizes bounds for cylcomatic complexity and what each suggests about a program.

\begin{table}[h]
\centering
\begin{tabular}{|l|l|}
\hline
0--5 & The routine is probably fine \\
\hline
6--10 & Consider simplifying the routine \\
\hline
10+ & Split logic into separate routines \\
\hline
\end{tabular}
\caption{Significance of cyclomatic complexity.}
\label{Table:Complexity}
\end{table}

The LCOM of a class from object oriented programming (OOP) is a measurement for how cohesive attributes and methods of classes are together. Low LCOM is desirable, and high LCOM suggests that classes should refactored because they are not modular. There are multiple versions of LCOM, and we use LCOM2, which is defined by the fraction,

\compact{
\begin{equation}
1 - \frac{\sum_{j=1}^{a}m(A_j)}{m \times a},
\end{equation}
}

where,

\begin{itemize}
\item $m$ is the number of methods in a class,
\item $a$ is the number of attributes in a class,
\item $A_k$ is the $k$th attribute of a class,
\item $m(A_k)$ is the number of methods that access attribute $A_k$.
\end{itemize}

Note that LCOM is calculated per class, and, therefore, we report the average across all classes.

A desirable quality of software is that is follows SOLID principles \autocite{Martin2017}. A detailed discussion of SOLID principles is out of the scope of this paper, however, we note that these qualities of software make programs loosely coupled. For example, we implemented a POJO for ant tour construction that does not depend on the Hadoop MapReduce framework in its imports. This class is loosely coupled, or, not tightly coupled, because editing our software that does depend on the Hadoop framework does not affect this POJO. Our software using the Hadoop framework effectively depends on the interface of this POJO that is core business logic. If we desired, we could swap usage of Hadoop with other frameworks like Spark or MPI without affecting this POJO that encapsulates ant tours, and methods with direct access to them. In other words, this makes our software, easy to change (ETC) \autocite{Thomas2020}.

\begin{table}[h]
\centering
\begin{tabular}{|c|l|}
\hline
S & Single responsibility principle (SRP) \\
\hline
O & Open-closed principle (OCP) \\
\hline
L & Liskov substitution principle (LSP) \\
\hline
I & Interface segregation principle (ISP) \\
\hline
D & Dependency inversion principle (DIP) \\
\hline
\end{tabular}
\caption{SOLID principles of software architecture.}
\end{table}

The design of our software is motivated to be, for example, easy to change, loosely coupled, and follow SOLID principles. However, these are qualitative properties, not quantitative ones. The quantitative measurements that we provide for our software, namely, LoC, lines of boilerplate code, cyclomatic complexity, and LCOM, are software engineering metrics that offer indications for the quality of our program. While they cannot capture design intent directly, they provide complementary evidence for maintainability, modularity, and overall program health.

\begin{table}[h]
\centering
\begin{tabular}{|c|c|}
\hline
LoC & 372 \\
\hline
Boilerplate & 90 \\
\hline
Cyclomatic complexity & 2 \\
\hline
LCOM & 0.18 \\
\hline
\end{tabular}
\caption{Software engineering metrics from our ant colony optimization program.}
\label{Table:SoftwareMetrics}
\end{table}

%% file: body/discussion/interpretation.tex
\subsection{Interpretation of Results}\label{Discussion:Interpretation}

The box plots of shortest distances from runs of parameter tuples in \autoref{Figure:BoxplotsExploration}, and \autoref{Figure:TableExploration}, show that many runs have similar performance in the \textit{Exploration} stage. Yet we rank parameter tuples by their mean shortest distance and select only the top 5 for the \textit{Exploitation} stage. In our \textit{Exploration} stage we limit computation to 3 runs of 10 iterations for ACO convergence with each of the 48 parameter tuples.

Other simulations of ACO applied to TSP show that the objective value of ACO exponentially decreases across iterations of convergence \autocite{Tsai2023}, and this suggests we do not need many iterations to get a sense of performance. However, 3 runs per parameter tuple is a small dataset for statistical analysis. Furthermore, 48 samples from a Sobol sequence offers a limited representation of four-dimensional parameter space. With a larger computation budget, we could generate more data for the purpose of performance ranking with greater statistical validity.

A similar interpretation can be made for our data generation methodology in the \textit{Exploitation} stage. We add more iterations for convergence because we are interested in tail end performance of ACO, yet 10 data points per parameter tuple is still a small dataset for each parameter tuple. Again, a higher computation budget would allow us to collect more data for more valid statistical analysis.

Beyond our data generation methodology, our use of parametric probability distributions is not altogether well-fitting. None our runs from parameter tuples in fact reach the optimum solution of TSPLIB berlin52. Therefore, we use a workaround by fitting continuous distributions, namely, Normal, Log-normal, Gamma, and Weibull, and estimate the probability of reaching the optimum solution by the CDF. This workaround risks bias in our models of the true distributions of our data, especially if MLE does not fit our data well.

If our runs with parameter tuples did reach the optimum solution of berlin52, then we could directly model our data with Binomial or Binomial-Beta distributions without depending on assumptions about the shape of the underlying continuous random variable. In that case, the conjugacy between the Binomial likelihood and the Beta prior would simplify posterior inference, allowing us to express uncertainty about success probabilities analytically. Moreover, because these distributions belong to the exponential family, we could exploit sufficient statistics, such as the count of successful runs, to summarize all information relevant to the posterior.

Furthermore, given results that we could model by the Binomial distribution, then could perform analytical uncertainty quantification of the posterior by the Beta distribution in closed-form. This offers us Bayesian credible intervals for the probability of reaching the optimum shortest path. In other words, we would not limited to methods such as numerical bootstrapping for uncertainty quantification in our \textit{Evaluation} stage. Our use of bootstrap sampling from our \textit{Evaluation} stage is an extension from our use of MLE to fit continuous distributions to our data.

In \autoref{Figure:QQPlotsExploitation} in \autoref{Appendix:Results} we see disappointing results from our Q-Q plots. Only \autoref{Figure:QQPlots11} from \autoref{Results:Exploitation} offers decent Q-Q plots from fitting MLE for runs of parameter tuple with index 11. We note again, however, that the size of our datasets with only 10 points from our \textit{Exploitation} stage could be giving us biased estimates for the success rate and reliability of reaching the optimum solution. This limitation suggests that a larger number of runs or additional replications could be necessary to obtain more stable and representative estimates of model fit.

Generally, the results of variance and stability from our case study could be improved. At the start of our \textit{Exploitation} stage, we see that runs of parameter tuples with low variance, consistent solutions do not come close to the optimum solution in \autoref{Figure:BoxplotsExploration}. At the same time, runs of parameters tuples with a high variance of reaching the optimum shortest distance have both the best and worse performance. Therefore, the parameter tuples that give us the greatest chance of reaching the optimum solution are unstable as shown in \autoref{Figure:Bootstrap11}, and this is undesirable.

\autoref{Table:ExploitationTuples} shows the values of parameter tuples used in the \textit{Exploitation} stage of our case study. Recall that parameter tuple with index 11 offered the best performance, that is, came closest to the optimum shortest path of berlin52 though with unstable performance. We see that index 11 has the highest $\alpha$, lowest $\beta$, lowest $\rho$, and second lowest number of ants. This parameter tuple suggests that it is benefiting from a high exploration ($\alpha$) to exploitation ($\beta$) ratio, relative to our other parameter tuples. In other words, index 11 relies on a posteriori pheromone information, rather than a priori edge distance (weight) information. This strategy avoid premature convergence.

Furthermore, the pheromone evaporation rate ($\rho$) of this strategy is low. This further supports that this strategy is benefiting from its a posteriori beliefs about optimal ant tour construction. Lastly, we note that its low number of ants could also benefit from an exploration-heavy search strategy. That is because for each iteration of ACO convergence, small steps are take toward the optimum solution, and this can avoid premature convergence. The values of parameter tuple with index 11 give us hints about why it was able to reach results near the optimum solution, as compared to other parameter tuples.

\begin{table}[h]
\centering
\begin{tabular}{|c|c|c|c|c|}
\hline
index & $\alpha$ & $\beta$ & $\rho$ & ants \\
\hline
11 & 1.837 & 3.141 & 0.201 & 155 \\
\hline
17 & 1.526 & 4.193 & 0.626 & 247 \\
\hline
31 & 1.387 & 3.376 & 0.499 & 185 \\
\hline
35 & 1.387 & 3.589 & 0.485 & 148 \\
\hline
43 & 1.009 & 4.984 & 0.281 & 244 \\
\hline
\end{tabular}
\caption{Parameter tuples from our \textit{Exploitation} stage in the EEE framework.}
\label{Table:ExploitationTuples}
\end{table}

In summary, we note that computation budget, data samples size, and biased estimation of successfully reaching the optimum shortest path for berlin52 are factors that interpretation of our results hinge on. Though, given the resources available for our case study, our estimations for ACO successfully solving TSP with TSPLIB berlin52 give satisfying results. Intuitively, our results are plausible.

The results from our software engineering metrics are mixed. We achieve strong metrics for cyclomatic complexity and LCOM as seen in \autoref{Table:SoftwareMetrics} from \autoref{Results:SoftwareEngineering}. Scores of 2 and 0.18 for cyclomatic complexity and LCOM, respectively, indicate that our software is readable (low cyclomatic complexity) and modular (low LCOM). SOLID principles design for software is not trivial. For example, our POJO that encapsulates ant tours could have more easily not been a POJO. The difference depends on object oriented programming class dependencies, and that comes down to decisions in software design and architecture. Therefore, we attribute our good scores for cyclomatic complexity and LCOM to our ability at successful software engineering.

On the other hand, software where boilerplate takes nearly 25\% of the LoC, as seen in \autoref{Table:SoftwareMetrics} from \autoref{Results:SoftwareEngineering}, is not desirable. The Hadoop MapReduce framework is highly opinionated, and this means that features and extensions of the framework are built on rigid assumptions of its paradigm of computation. Therefore, the Hadoop MapReduce framework naturally has a high ratio of boilerplate to business logic. We attribute the undesirable, high ratio of boilerplate to business logic as a reflection of the Hadoop MapReduce framework, rather our aptitude at software engineering.

We also reflect on the runtime of Hadoop. Recall from \autoref{Results:Exploration} and \autoref{Results:Exploitation} that runtime for one run of 10 iterations for ACO convergence took 4 minutes. This time constraint alone is a major barrier to executing wider exploration and deeper exploitation of the parameter space. The Hadoop MapReduce framework reads from and writes to disk between $\operatorname{Mapper}$ and $\operatorname{Reducer}$ operators, and between Hadoop MapReduce jobs, that is, between iterations of ACO convergence. Frequent disk access makes runtime slow. For the purpose of faster runtime of ACO, and to keep parallel and distributed execution, we should implement ACO with Spark or MPI frameworks. Therefore, in addition to its poor ratio of boilerplate to business logic, the Hadoop MapReduce framework can have slow runtime caused by its frequent disk access.

%% file: body/discussion/strengths_limitations.tex
\subsection{Strengths and Limitations}\label{Discussion:StrengthsLimitations}

The Exploration--Exploitation--Evaluation framework gives us a flexible means for calculating the success rate and reliability of a metaheuristic algorithm solving a combinatorial optimization problem. It performs a shallow search of the parameter space to avoid high computation cost, a deep search that enables better insights from data analysis like MLE, and numerical resampling to measure reliability. The use of MLE and resampling like the bootstrap method is robust to cases, such as our case study, where the optimum solution is never in fact reached and so we cannot directly model outcomes with a Binomial or Binomial-Beta distribution nor analytically solve for probabilities or express uncertainties in closed-form.

However, the ability of our EEE framework to work around zero empirical successes comes at the cost of bias with the distributions that we use to model our data. In other words, the EEE framework is never blocked from calculating the probability of successfully reaching the optimum solution. Additionally, its ability to always complete uncertainty quantification gives us an indication as to whether our results are trustworthy due to their reliability. Therefore, the strength of our EEE framework is that it can always give a numerical answer to the success rate of a metaheuristic algorithm reaching the optimum solution of a combinatorial optimization problem. However, its limitation is that we may not always want to trust that answer.

A major strength of our case study is that we successfully estimated the probability and confidence intervals of reaching the optimum solution for TSPLIB berlin52. However, a limitation is that we did not have a high computation budget, with our most limiting constraint being the use of the Hadoop MapReduce framework which offers slow iterations for ACO convergence, as previously discussed. However, consider that our choice of ACO algorithm may have also been a limitation in our results.

We implemented the more simple ant system (AS), but, ant colony system (ACS) is known to give better performance \autocite{Tsai2023}. In AS, the transition probability is computed similar to roulette wheel selection which is not an efficient method of search, and comes with higher computation cost. On the other hand, the transition rule of ACS more deterministically chooses the optimal transition for better search efficiency, and this lowers computation cost. The strength of using AS is that is offers a baseline estimate for the success of ACO to solve TSP, however, its limitation is that ACS is known to have better performance.

We acknowledge again that our use of the Hadoop framework creates a low-ceiling constraint on our computation budget. To maintain distributed, parallel execution and for faster runtime speed, we should consider using the Spark or MPI frameworks. These frameworks do not consistently access disk as part of their computation paradigm. They keep data in memory where access for manipulation is much faster. Faster computation allows for wider exploration and deeper exploitation of the parameter space for ACO. In summary of computation costs, the strength of the Hadoop MapReduce framework is that it works and gives us distributed and parallel execution, however, its limitation is that frequent disk access is a bottleneck.

Lastly, we comment on the strengths and limitations of the Hadoop MapReduce framework from a software engineering perspective. The Hadoop framework allows us to write parallel programs that encapsulates distribution and networking of work across a cluster of nodes. This is in contrast, for example, to MPI that requires us to take greater control of networking between machines. This encapsulation of networking lets us focus on business logic in the Hadoop MapReduce computation paradigm. However, as we have seen in table \autoref{Table:SoftwareMetrics} from \autoref{Results:SoftwareEngineering}, this does not absolve Hadoop from its high boilerplate to business logic ratio.

Additionally, the expressiveness of object oriented programming is not lost when we use Hadoop. We showed this with our good cyclomatic complexity and LCOM. However, the MapReduce framework paradigm can be rigid, making it difficult in some cases to convert programming requirements to software that fit the $\operatorname{Mapper}$ and then $\operatorname{Reducer}$ operators workflow. In summary, the strengths of the Hadoop MapReduce framework is that it allows software engineers to write parallel programs with computation distributed across multiple machines, and it does not limit the expressiveness of Java and object oriented programming. However, its ratio of boilerplate to business logic is high, and this is undesirable. Furthermore, the MapReduce framework is not always intuitive for software engineering.

%% file: body/discussion/future.tex
\subsection{Future Directions}

Our Exploration--Exploitation--Evaluation framework offers a straightforward methodology to measure the success rate and reliability of using metaheuristic algorithms to solve combinatorial optimization problems. With a higher computation budget for our case study we could collect more data and better understand how limiting the assumptions of MLE, the CDF, and the bootstrap method are for estimation of the probability of success with confidence intervals. For example, greater data acquisition, such as more parameter tuples, more runs per parameter tuple, and more iterations per run, would indicate if our methodology is what limits the ability of ACO to solve TSP, or alternatively if the ACO algorithm itself is a more likely limiting factor. Future directions of research include repetition of a case study with greater data acquisition.

Also consider that we used the ant system algorithm rather than the ant colony system algorithm for our implementation of ant colony optimization. As we have discussed in \autoref{Discussion:StrengthsLimitations}, ACS has better performance than AS because it more deterministically selects the optimal transition during ant tour construction. It is possible, therefore, that ACS is able to achieve reaching the optimum solution with small sample sizes whereas AS is not. This means we could expand the EEE framework to include Binomial and Binomial-Beta distribution modeling given runs of ACO that in fact reach the optimum shortest distance of TSP. Another future direction of research is to apply a case study of ACO to TSP that uses ACS as its underlying algorithm.

We have considered the trustworthy expressiveness of directly modeling a success rate from data that in fact has empirically successful outcomes. Assuming we achieve the optimum shortest distance in a case study, which could be seen from greater data acquisition or metaheuristic algorithms as discussed above, then we could update documentation of the EEE framework to include directly modeling the success rate and reliability from empirical data. This would create a decision point in our framework, which offers more trustworthy end results when the option to apply this option is available. A future direction of research is to document this approach so that it is easy to apply in future case studies.

Another promising direction of future research is the use of the Spark of MPI frameworks for parallel, distributed execution of metaheuristic algorithms to solve combinatorial optimization problems. The Hadoop MapReduce framework is too slow for high iteration algorithms since disk access is built-in to its operations. The Spark and MPI frameworks do keep data in memory which offers much faster iterations and data manipulations for metaheuristic algorithms. Leveraging these in-memory frameworks could enable scalability to larger problem instances and also facilitate real-time or near-real-time optimization in dynamic environments.

Lastly, we consider another interesting direction of research for a case study. Metaheuristic algorithms have been applied to deep neural networks to find optimum architecture backbones that describe a combinatorial optimization problem. Additionally, metaheuristic algorithms have also been to deep learning for hyperparameter tuning which represents a continuous optimization problem. For the purpose of applying metaheuristic algorithms to optimize deep learning performance, metaheuristic algorithms beyond ant colony optimization could be researched in a case study that applies the EEE framework.

%% file: body/conclusion.tex
Currently, there is a lack of standardization for how to benchmark metaheuristic algorithms used to solve combinatorial optimization problems, and continuous optimization problems. for this purpose, in this paper we introduced the Exploration--Exploitation--Evaluation framework that provides estimates of the success rate and reliability of metaheuristic algorithms to solve COPs and continuous optimization problems. We also report the results of a case study that applies the EEE framework using ant colony optimization to solve the traveling salesman problem. Our case finds that the probability of ACO optimally solving the TSP for berlin52 is 1/40 across one run, and improves to 1/5 across ten runs. Interesting future directions of research include applying the EEE framework to deep learning research by applying metaheuristic algorithms to deep neural network architectures that fit the description of a combinatorial optimization problem, and hyperparameter tuning that fit the description of a continuous optimization problem.

%% file: appendix/optimization_metaheuristics.tex
\section{Optimization \& Metaheuristics}\label{Appendix:Meta}

Optimization problems are composed of an objective function, constraint, and solution. The most widely known problems are combinatorial and continuous optimization problems. Combinatorial problems include one-max, 0-1 knapsack, binary to decimal, and traveling salesman problems. Analogously, continuous optimization problems include single-objective and multi-objective problems.

A formal definition is needed to describe the characteristics of a problem in order to develop a search algorithm to solve it.

\textbf{Optimization problem $\mathbb{P}$:}

An optimization problem $\mathbb{P}$ is to find the optimal value, possibly subject to some constraints, out of all possible solutions \autocite{Tsai2023}.

\compact{
\begin{equation}
\opt_{s\in\mathbf{A}}f(s), \; \text{subject to} \; \forall c_i \odot b_i, \; i=1,2,\ldots,m,
\end{equation}
}

where,

\begin{itemize}
\item $\opt$ is either $\min$ or $\max$,
\item $s$ is a candidate solution,
\item $\mathbf{A}$ and $\mathbf{B}$ are the domain and codomain of the problem $\mathbb{P}$, namely, $\mathbf{A}$ is the set of all possible solutions, and $\mathbf{B}$ is the set of all possible outcomes of the objective function,
\item $f(s):\mathbf{A}\to\mathbf{B}$ is the objective function,
\item $c_i(s) \odot b_i$ is the $i$th constraint, and
\item $\odot$ is $<, >, =, \leq, \text{or} \geq$.
\end{itemize}

The optimal solution and optimal value of the optimization problem $\mathbb{P}$ are defined as follows.

\textbf{Optimal solution $s^\ast$:}

The optimal solution is a solution, out of all possible candidate solutions of the optimization problem $\mathbb{P}$, that gives the optimal value \autocite{Tsai2023}.

\compact{
\begin{equation}
f(s^\ast) = \opt_{s\in\mathbf{A}}f(s), \; \forall c_i(s) \odot b_i, \; i=1,2,\ldots,m.
\end{equation}
}

\textbf{Optimal value $f^\ast$:}

If the optimal solution $s^\ast$ for the problem $\mathbb{P}$ exists, then the optimal value $f^\ast$ is defined as follows \autocite{Tsai2023}.

\compact{
\begin{equation}
f^\ast=f(s^\ast).
\end{equation}
}

Optimization problems can be classified into two categories based on the variable type of the solution space, namely, solutions encoded as discrete variables and solutions encoded as continuous variables. The former is referred to as the combinatorial optimization problem (COP), and the latter is referred to as the continuous optimization problem. For a COP, we are looking for the best solution from a finite set, for example, a set of integers, a permutation, or a graph structure. For a continuous optimization problem, we are looking for a set of real numbers.

The \textit{traveling salesman problem} (TSP) is a COP that is defined as follows. Given $n$ cities and the distance between each pair of cities, seek a tour (a closed path) with minimum distance that visits each city in sequence once and only once and returns directly to the first city. Formally, TSP is defined as follows \autocite{Tsai2023}.

\compact{
\begin{equation}
\min_{s \in c_\pi} f(s) = \left[ \sum_{i=1}^{n-1}d(c_{\pi(i)}, c_{\pi(i+1)}) \right] + d(c_{\pi(n)}, c_{\pi(1)}),
\end{equation}
}

where $c_\pi = \{ <c_{\pi(1)},c_{\pi(2)},\ldots,c_{\pi(n)}> \}$, that is, all permutations of the $n$ cities.

Compared to classical exhaustive search (ES) and hill climbing (HC) algorithms, a metaheuristic algorithm will neither check all the candidate solutions of an optimization problem like ES nor fall into a local optimum at early iterations as easily as HC. Since metaheuristics have been widely used and discussed since the 1960s or earlier, they can be regarded as a distinguished school of contemporary search methods. Several engineering and commercial applications show that metaheuristic algorithms are able to find good results in a reasonable time. This is one of the factors the increase the development of metaheuristic algorithms. For example, ant colony optimization algorithms are empirically useful for solving combinatorial optimization problems \autocite{Tsai2023}.

Since 2010 or earlier, some groups have attempted to apply metaheuristic algorithms in high-performance computing (HPC) environments. Some distributed or parallel metaheuristics are not only able to provide end results more quickly, they can also find better results than a single machine because the parallel computing mechanism leads them to increase the search diversity during the convergence process. Distributed, parallel, and clouding computing environments (Hadoop, Spark, Microsoft Azure, Amazon EC2, and Google Compute Engine) offer an easy way to access distributed computing systems. However, in such a distributed computing system, communication cost is critical: if computation is not large enough to offset it, metaheuristics may run slower than on a single machine. This is a noteworthy research trend.

The appearance of metaheuristic algorithms has come with methods to classify them including the following: (1) nature vs. non-nature inspired, (2) dynamic vs. static objective function, (3) one vs. various neighborhood structures, (4) memory vs. memoryless, (5) with vs. without local search, (6) population-based vs. single-solution-based search. Many metaheuristic algorithms exist and most of them are described in different ways. The unified framework for metaheuristics (UFM) explains the basic idea of metaheuristic algorithms.

UFM consists of give main operators: initialization, transition, evaluation, determination, and output. In this framework $\mathbb{I}$ denotes the input dataset, $s$ denotes the current solution, $v$ denotes the candidate solution, $f_s$ denotes the objective value of $s$, and $f_v$ denotes the objective value of $v$. Also, $s$ and $v$ can denote either a single solution or a set of solutions, where each solution has $n$ elements or is an $n$-tuple. The role of each operator is as follows \autocite{Tsai2023}.

\begin{itemize}
\item \textbf{Initialization:} Plays the role of reading the input file, initializing all the parameters of a metaheuristic algorithm, and determining the initial solution, which is normally based on a random process.
\item \textbf{Transition:} Plays the role of varying the search directions, such as perturbing a portion of the subsolutions of the current solution to generate a new candidate solution.
\item \textbf{Evaluation:} Responsible for measuring the quality of solutions, such as calculating the objective value of each solution to be used by the determination operator to distinguish the quality of all solutions.
\item \textbf{Determination:} Plays the role of deciding the search directions by using information the evaluation operator provides during the convergence process. The performance of metaheuristic algorithms largely depend on the performance of this operator. A good search strategy for this operator will make it possible for the metaheuristic algorithm to find a better solution faster or to avoid falling into a local optima at early iterations.
\item \textbf{Output:} This operator can be either simple or complex depending on how much information we want to display for the metaheuristic algorithm. Output can range from showing only the final result to showing the full convergence trajectory for deeper performance insights.
\end{itemize}

\begin{algorithm}
\caption{Unified Framework for Metaheuristics (UFM) \autocite{Tsai2023}}
\begin{algorithmic}[1]
    \State $s \gets \operatorname{Initialization}(\mathbb{I})$
    \State $f_s \gets \operatorname{Evaluation}(s)$
    \While{the termination criterion is not met}
        \State $v \gets \operatorname{Transition}(s)$
        \State $f_v \gets \operatorname{Evaluation}(v)$
        \State $s, f_s \gets \operatorname{Determination}(s, v, f_s, f_v)$
    \EndWhile
    \State $\operatorname{Output} s$
\end{algorithmic}
\end{algorithm}

The initialization and output operators will be performed only once, and the transition, evaluation, and determination operators will be performed repeatedly for a certain number of times. The transition, evaluation, and determination operators can be regarded as the core of the algorithm, which is aimed at searching for the solution. The initialization operators can be regarded as the interface between the input dataset and the metaheuristic algorithm. Moreover, the output operator can be regarded as the interface between the solution and the human being or the other algorithms, including metaheuristics.

The \textit{ant colony optimization} (ACO) algorithm is a popular algorithm of swarm intelligence. The search behavior of ACO is inspired by the behavior of ant colonies. It is not a single algorithm but a collection of ant based search algorithms. For this kind of metaheuristic algorithm, the routing path (tour) of an ant represents a solution to the problem. Each ant constructs a solution step by step by using the distance information and the searched information saved in the pheromone table.

ACO and its variants are typically very useful in solving optimization problems where the structure of solutions is a sequence of subsolutions, such as TSP. Since ACO is also a population-based metaheuristic algorithm, it will create a set of solutions $s = \{ s_1, s_2, \ldots, s_m \}$ in the very beginning, where $s$ denotes a set of routing paths of ants, $s_i = \{ s_{i,1}, s_{i,2}, \ldots, s_{i,n} \}$ denotes a routing path of an ant (a solution), and $s_{i,j}$ denotes the $j$th subsolution of the $i$th ant. Note that $m$ and $n$ denote the number of ants in the population and the number of subsolutions in an ant, respectively.

ACO-based algorithm usually use both distances between nodes and the pheromone table together to determine the node for the next move. The higher the pheromone value of an edge, the higher the number of ants traveling through it. Analogously, the lower the pheromone value of an edge, the lower the number of ants traveling through it. The distances between nodes can be regarded as the local information seen by an ant that guides the ant to choose an edge that is as short as possible. The pheromone on the edges can be regarded as the global information contributed by all the ants since the very beginning that guides an ant to choose the edge that has been traveled by a larger number of ants. From these observations, it can be seen that ACO takes into account the distance and pheromon information at the same time in the determination of search directions.

\begin{algorithm}
\caption{Ant Colony Optimization \autocite{Tsai2023}}\label{Algorithm:ACO}
\begin{algorithmic}[1]
    \State Initialize the pheromone values $\tau$ for all the edges
    \While{the termination criterion is not met}
        \State $s \gets \operatorname{SolutionConstruction}(\tau, \operatorname{dist})$
        \State $f \gets \operatorname{Evaluation(s)}$
        \State $\tau \gets \operatorname{PheromoneUpdate}(s,f)$
    \EndWhile
    \State $\operatorname{Output} s$
\end{algorithmic}
\end{algorithm}

The operator $\operatorname{SolutionConstruction}()$ shown in \autoref{Algorithm:ACO} will first randomly put each ant $a^k$on a node $c_i$ as the starting point of each ant, and it will then compute the transition probability from the node $c_i$ to the other nodes that have not been visited. The transition probability is computed as follows \autocite{Tsai2023}.

\compact{
\begin{equation}\label{Equation:Transition}
s_{k,i} = p_{i,j}^{k} = \frac{\tau_{i,j}^{\alpha}\eta_{i,j}^{\beta}}{\sum_{g\in\mathbb{N}_i^k}\tau_{i,g}^{\alpha}\eta_{i,g}^{\beta}},
\end{equation}
}

where $s_{k,i}$ denotes the $i$th subsolution of the $k$th ant, $p_{i,j}^{k}$ denotes the transition probability of the $k$th ant at node $c_i$ choosing node $c_j$ as the next node, $\tau_{i,j}$ and $\tau_{i,g}$ denote the pheromone values, $\eta_{i,j}$ and $\eta_{i,g}$ denote the heuristic values, $\alpha$ and $\beta$ denote the importance of $\tau_{i,j}$ and $\eta_{i,j}$, and $\mathbb{N}_i^k$ denotes the set of candidate subsolutions that can be chosen by ant $k$ at node $i$.

The value $\eta_{i,j}$ is called the visibility, which is defined as $1/d_{i,j}$ where $d_{i,j}$ is the distance between node $c_i$ and node $c_j$. Nodes are chosen step by step, and at each step, the number of nodes that can be chosen will be decreased, and the transition probability will also be recomputed for choosing the next node. A complete solution will be constructed once all the nodes have been chosen and visited.

As seen in \autoref{Algorithm:ACO}, after all the solutions are constructed by all the ants, the $\operatorname{PheromoneUpdate}()$ operator is performed to update the pheromone values of all the edges. This operator will first compute the pheromone value of each ant as follows \autocite{Tsai2023}.

\compact{
\begin{equation}
\Delta\tau_{i,j}^k = \begin{cases}
\dfrac{Q}{L_k} & \text{if the $k$th ant uses edge $e_{i,j}$}, \\
0 & \text{otherwise},
\end{cases}
\end{equation}
}

where $Q$ is a constant and $L_k$ is the tour length of the $k$th ant. That is, the pheromone value of each ant $\Delta\tau_{i,j}^k$ is inversely proportional to the quality of the solution. Next, given the values $\Delta\tau_{i,j}^k$ of each ant, this operator will compute the accumulated pheromone value for each edge $e_{i,j}$ as follows \autocite{Tsai2023}.

\compact{
\begin{equation}
\Delta\tau_{i,j} = \sum_{k=1}^{m}\Delta\tau_{i,j}^k,
\end{equation}
}

where $m$ is the number of ants. Finally, this operator will update the pheromone value of each edge as follows \autocite{Tsai2023}.

\compact{
\begin{equation}
\tau_{i,j} = \rho \times \tau_{i,j} + \Delta\tau_{i,j},
\end{equation}
}

where $\rho$ is a weight in the range $[0,1)$.

Most ACO-based algorithms use \autoref{Equation:Transition}, which is computationally expensive, to compute the transition probability. Therefore, they can be easily accelerated by reducing the computation time of the transition probability method. Parallel computing methods provide an efficient way to accelerate the response time of ACO-based algorithms. Different parallel computing architectures can be used, such as multi-core processors, graphical processing units (GPUs), and grid environments.

In summary, as mentioned before, since most ACO-based algorithms use the solution construction procedure to build a solution, they provide an effective approach for combinatorial optimization problems. As a result, most ACO-based algorithms are typically used to solve combinatorial optimization problems. This overview establishes the foundation for applying the principles of these metaheuristic algorithms within the context of optimization problems.

%% file: appendix/cluster.tex
\section{Hadoop Cluster in AWS EC2}\label{Appendix:Cluster}


Detailed setup instructions and screenshots for reproducing our Hadoop cluster in AWS EC2 implementation are available in the GitHub repository of this paper \autocite{Davis2025}. The instructions can be found from the root of our repository under $\operatorname{/cluster/readme.md}$. Note that for this paper we used $\operatorname{t3.xlarge}$ AWS EC2 instances, however, in the instructions $\operatorname{t2.xlarge}$ instances are used.

%% file: appendix/results.tex
\section{Additional Results}\label{Appendix:Results}

\begin{figure*}[p]
\centering
\includegraphics[width=\linewidth]{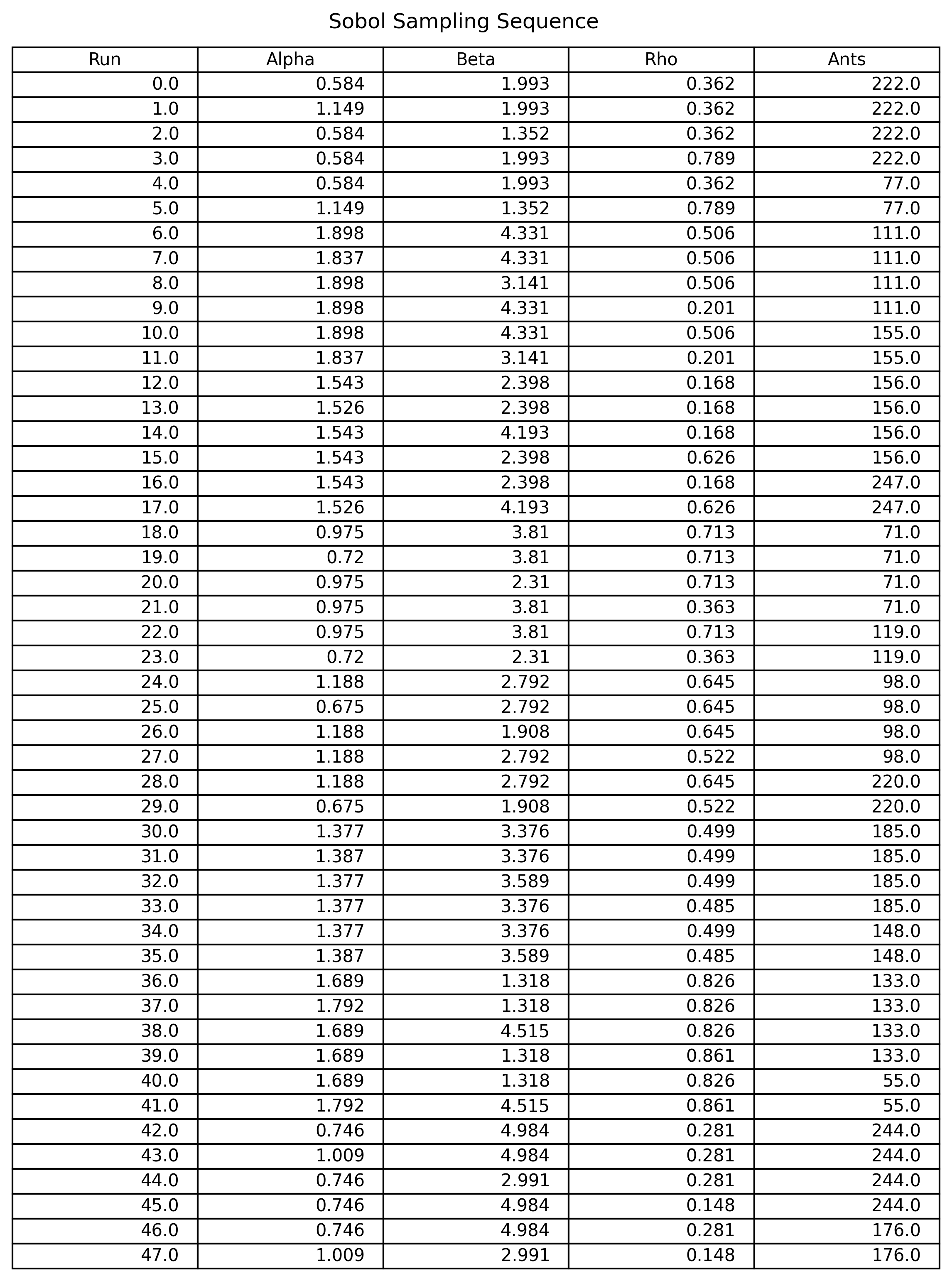}
\caption{Sobol sampling sequence for our EEE framework.}
\label{Figure:Sobol}
\end{figure*}

\begin{figure*}[p]
\centering
\includegraphics[width=0.15\linewidth]{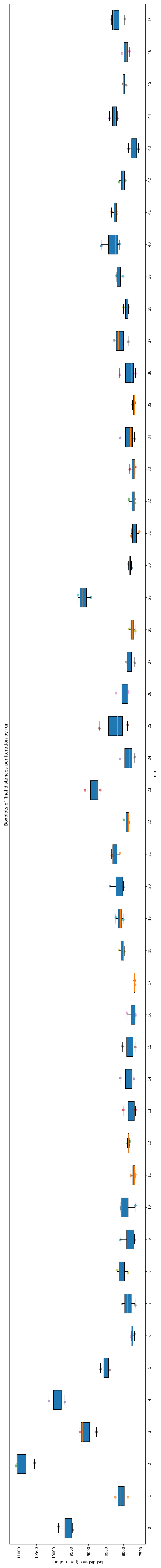}
\caption{Shortest distance box plots from the \textit{Exploration} stage in our EEE framework.}
\label{Figure:BoxplotsExploration}
\end{figure*}

\begin{figure*}[p]
\centering
\includegraphics[width=0.8\linewidth]{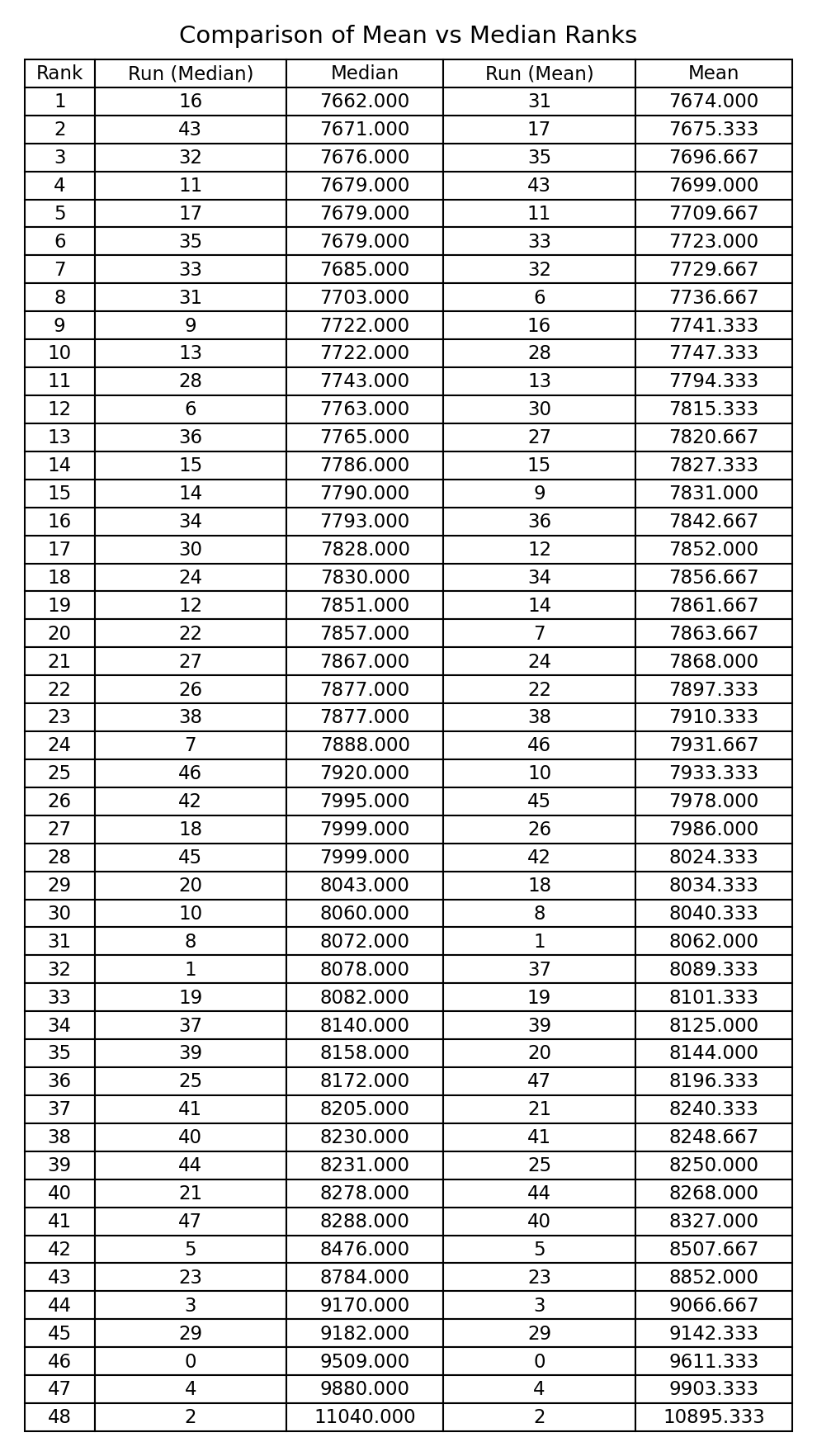}
\caption{Shortest distance rankings table from the \textit{Exploration} stage in our EEE framework.}
\label{Figure:TableExploration}
\end{figure*}

\begin{figure*}[p]
\centering
\includegraphics[width=\linewidth]{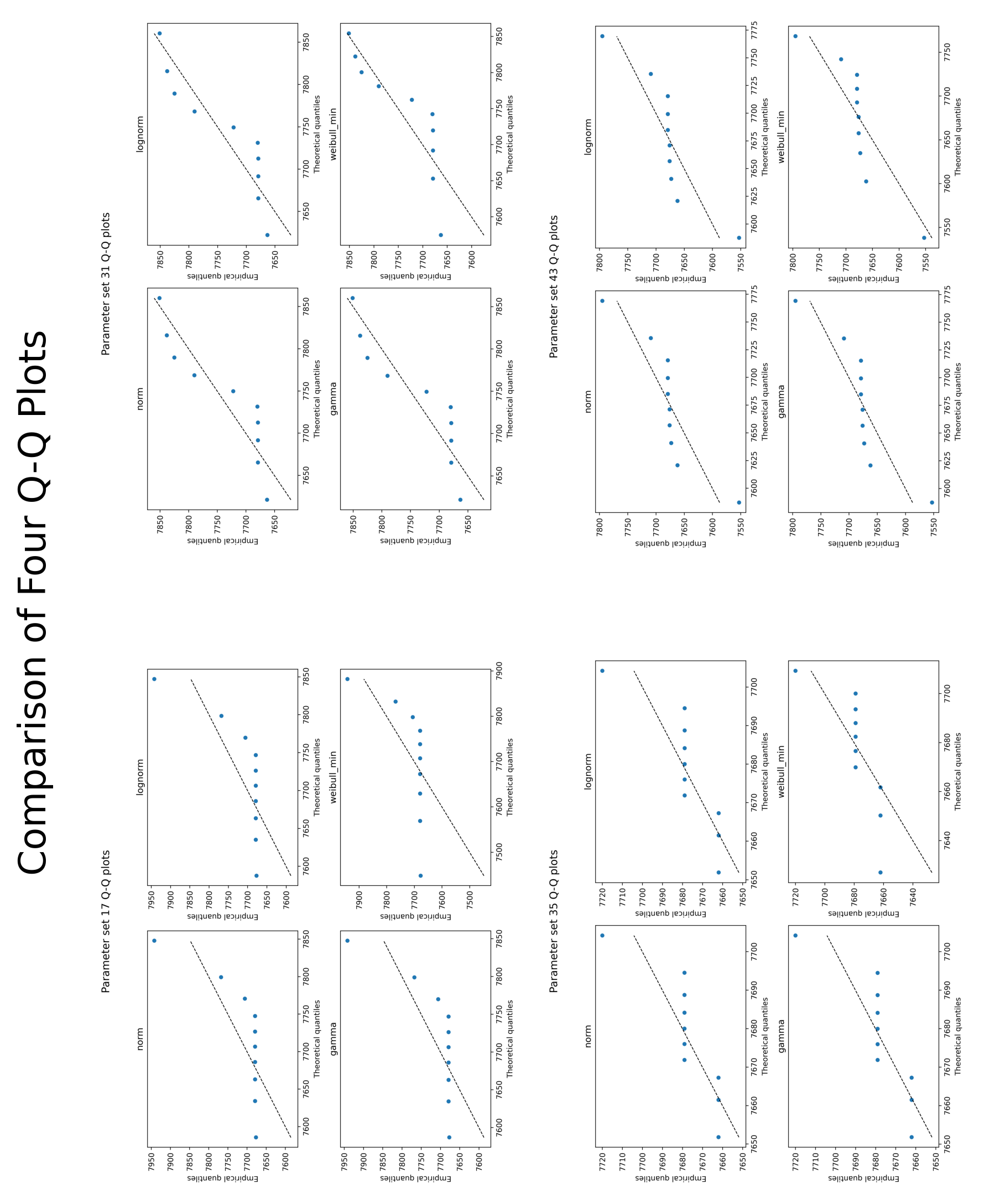}
\caption{Q-Q plots from runs in our \textit{Exploitation} stage of the EEE framework.}
\label{Figure:QQPlotsExploitation}
\end{figure*}

\begin{figure*}[p]
\centering
\includegraphics[width=0.5\linewidth]{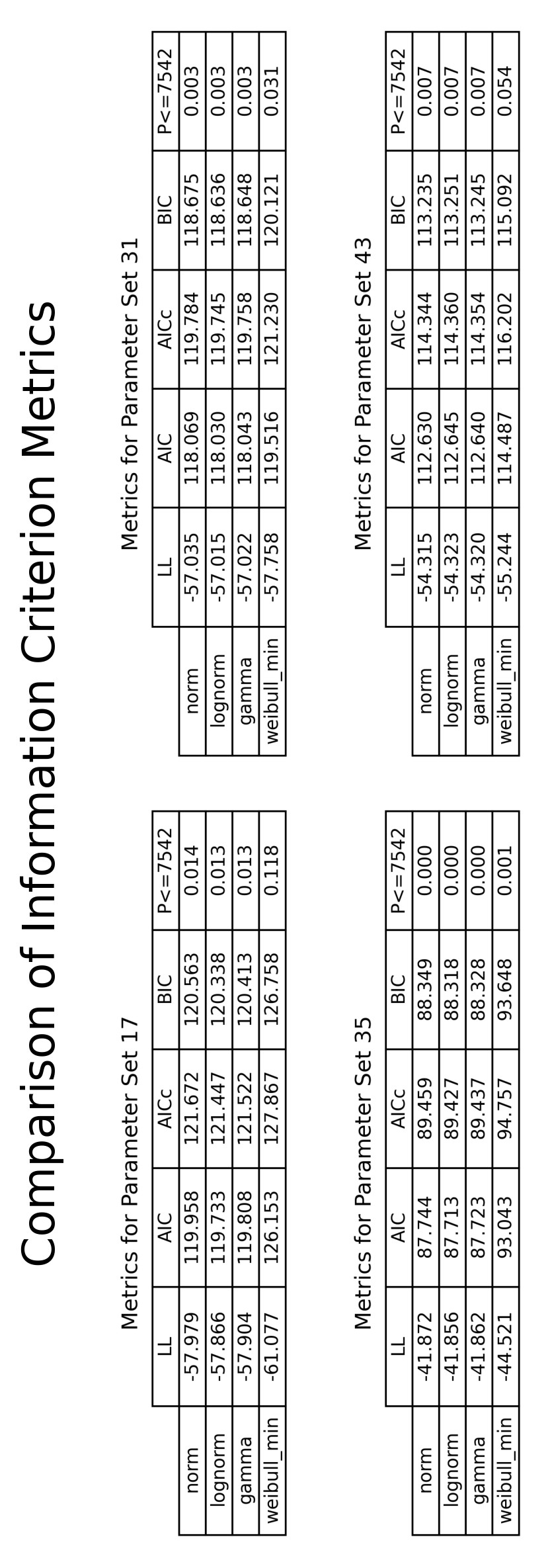}
\caption{Information criterion metrics from the \textit{Exploitation} stage in our EEE framework.}
\label{Figure:MetricsExploitation}
\end{figure*}

\begin{figure*}[p]
\centering
\includegraphics[width=0.33\linewidth]{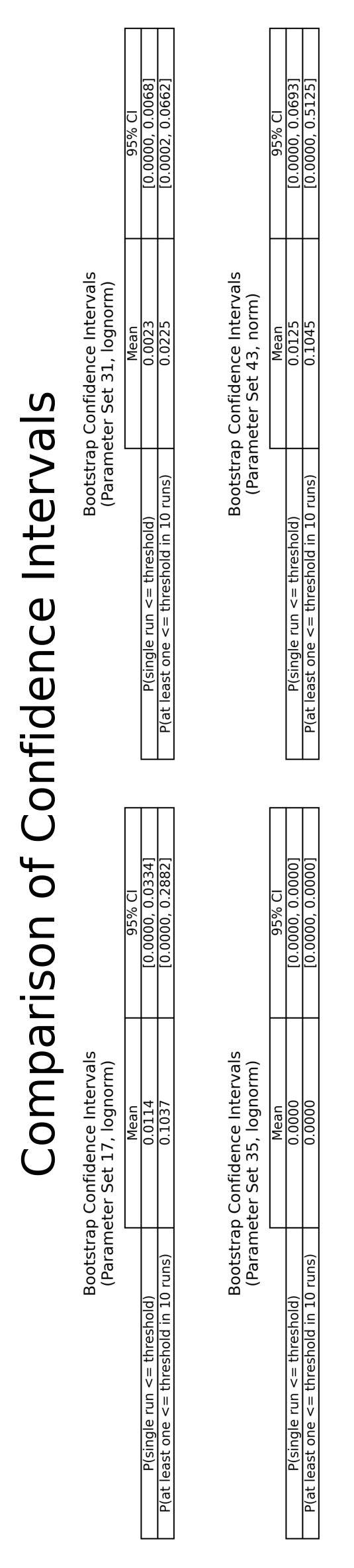}
\caption{Bootstrap confidence intervals from our \textit{Exploitation} stage of the EEE framework.}
\label{Figure:BootstrapEvaluation}
\end{figure*}